\documentclass{article}

\usepackage{arxiv}
\PassOptionsToPackage{numbers}{natbib}
\usepackage[utf8]{inputenc} 
\usepackage[T1]{fontenc}    
\usepackage{hyperref}       
\usepackage{url}            
\usepackage{booktabs}       
\usepackage{amsmath,amssymb,amsfonts}
\usepackage{nicefrac}       
\usepackage{microtype}      
\usepackage{lipsum}		
\usepackage{graphicx}
\usepackage{natbib}
\usepackage{doi}
\usepackage{tabularx}
\usepackage{textcomp}
\usepackage{hyperref} 
\usepackage{algorithmic}

\usepackage{CJKutf8} 
\usepackage{caption}
\usepackage{stackengine}

\title{Psycho Gundam: Electroencephalography based real-time robotic control system with deep learning}

\date{} 					

\author{ \href{https://orcid.org/0000-0003-0807-0217}{\includegraphics[scale=0.06]{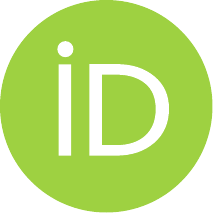}\hspace{1mm}Chi-Sheng Chen\thanks{Corresponding author} \thanks{These authors’ contributions are equal.}} \\
 Neuro Industry, Inc.\\
	\texttt{m50816m50816@gmail.com} \\
	\And
	\href{https://orcid.org/0009-0001-3572-5736}
 {\includegraphics[scale=0.06]{orcid.pdf}\hspace{1mm}Wei-Sheng Wang\thanks{These authors’ contributions are equal.} \thanks{This paper has been accepted by 2024 13th International Symposium on ACG Culture and Technology Studies (U-ACG 2024). More details at \protect\url{https://u-acg.com/archives/29015}.}} \\
  Tsukuba University\\
	\texttt{qwe789qwec@gmail.com} \\
}

\renewcommand{\shorttitle}{\textit{arXiv} Template}

\hypersetup{
pdftitle={A template for the arxiv style},
pdfsubject={cs.CV},
pdfauthor={Chi-Sheng ~Chen, Guan-Ying ~Chen, Dong ~Zhou, Dai-Shi ~Chen},
pdfkeywords={Deep learning, fine-grained visual classification, food category recognition, state space model, mamba},
}

\begin{document}
\maketitle

\begin{abstract}
	The Psycho Frame, a sophisticated system primarily used in Universal Century (U.C.) series mobile suits for NEWTYPE (\begin{CJK}{UTF8}{min}ニュータイプ\end{CJK}) pilots, has evolved as an integral component in harnessing the latent potential of mental energy. Its ability to amplify and resonate with the pilot’s psyche enables real-time mental control, creating unique applications such as psychomagnetic fields and sensory-based weaponry. This paper presents the development of a novel robotic control system inspired by the Psycho Frame \cite{gundam}, combining electroencephalography (EEG) and deep learning for real-time control of robotic systems. By capturing and interpreting brainwave data through EEG, the system extends human cognitive commands to robotic actions, reflecting the seamless synchronization of thought and machine, much like the Psyco Frame’s integration with a Newtype pilot’s mental faculties. This research demonstrates how modern AI techniques can expand the limits of human-machine interaction, potentially transcending traditional input methods and enabling a deeper, more intuitive control of complex robotic systems.
\end{abstract}


\section{Introduction}
The notion of using mental energy to control complex machinery has long been a key theme in science fiction, especially in the U.C. Gundam series. The Psycho Frame, a technology specifically designed for Newtype pilots, epitomizes this concept by enabling the pilot to control a mobile suit via mental synchronization, enhancing not only their combat effectiveness but also the machine's response to mental stimuli. This technology amplifies the pilot's mental power, creating psychomagnetic fields and enabling long-range, remote-controlled attacks. With the evolution of the Psycho Frame into the NT-D system, the pilot’s mental abilities are further augmented, allowing for advanced applications such as quantum manipulation and conceptual interference with physical systems.

Inspired by this futuristic vision, we explore the real-world implications of mental control using EEG-based robotic systems. Electroencephalography captures brainwave signals that are processed by deep learning algorithms to establish real-time control over robotic platforms \cite{an2023eeg}. This work builds upon the concept of the Psycho Frame to develop a control system where mental inputs can direct robot actions. Our system, drawing from advanced AI and neurotechnology, aims to emulate the seamless interaction between human thought and robotic behavior, as seen in the Gundam universe. The paper outlines the design, training, and deployment of this EEG-based control system, analyzing its potential for applications in fields such as prosthetics, robotics, and augmented human-machine interfaces.

The contributions of this work are stated as follows:

\section{Related work}
\label{sec:headings}


\subsection{EEG-based robotic control}
EEG-based robotic control has garnered significant attention in recent years, particularly in the fields of brain-computer interfaces (BCI) with prosthetic limb \cite{xu2022continuous}. Traditional BCIs leverage EEG signals to interpret neural activity for controlling external devices, such as robotic arms or wheelchairs. Early systems used pre-defined brainwave patterns, such as motor imagery or visual evoked potentials, to map user intentions to robotic actions \cite{wolpaw2002brain}. However, these systems faced limitations in terms of signal clarity, processing speed, and user adaptability.

Recent advancements in deep learning have significantly enhanced the both capabilities of EEG-based control systems and EEG encoders \cite{chen2024mind, chen2024necomimi}. Convolutional neural networks (CNNs) and recurrent neural networks (RNNs) have been applied to EEG signal classification with greater accuracy, enabling more precise and fluid control over robotic systems \cite{lawhern2018eegnet}. Moreover, transfer learning techniques have been utilized to reduce the amount of training data needed for each user, addressing the challenge of individual variability in brainwave patterns.

In parallel, research in neuroprosthetics has explored the integration of BCI systems with robotic limbs, allowing users to control prosthetic devices via neural signals. These systems have demonstrated promising results in enabling individuals with motor impairments to perform complex tasks (Brauchle et al., 2018). The combination of BCI and robotic control holds potential not only for assistive technologies but also for enhancing human-machine interaction in various fields, including teleoperation, gaming, and exoskeletons.

This work builds upon these developments, utilizing deep learning models to enhance the real-time processing of EEG signals for precise control of robotic systems. By drawing inspiration from science fiction’s portrayal of mental-robotic synchronization, this approach seeks to push the boundaries of intuitive, thought-based robotic control.

\section{Methodology}
\begin{figure}
    \centering
    \includegraphics[width=1\linewidth]{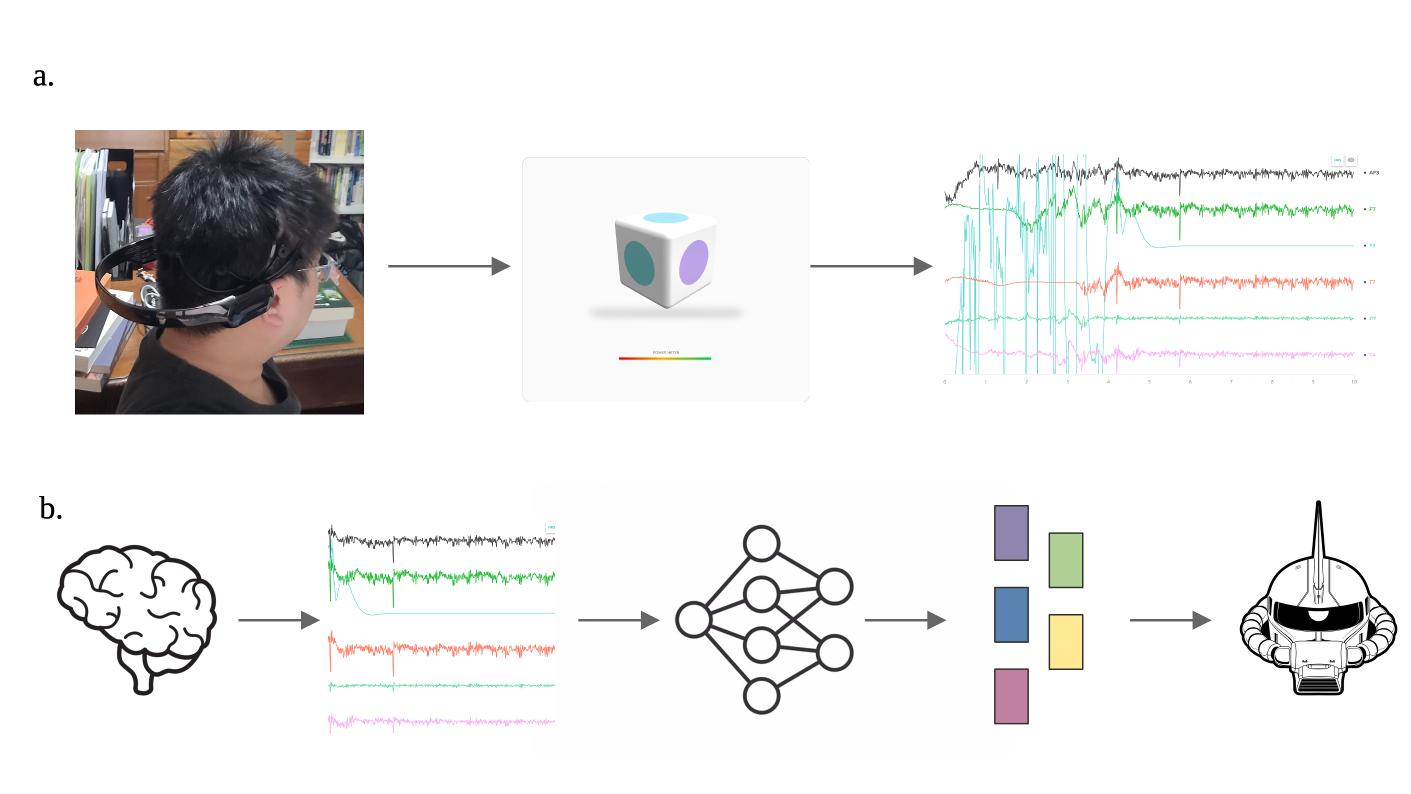}
    \caption{BCI-based Robotic Control System for Motor Imagery (MI) Training and Signal Decoding. (a) In the MI training stage, the user visualizes pushing or pulling a virtual box in five directions by using the action of the cockpit to imagine: forward, backward, left, right, and upward. This generates five distinct types of MI signals, and the raw EEG signals are recorded for training purposes. (b) A neural network is trained to decode these EEG signals into five control commands, which are transmitted to the robot. The robot then executes actions based on the one-hot decoded commands, enabling real-time control through mental visualization.}
    \label{fig: bci_abs}
\end{figure}
\subsection{Overview}
This section describes the methodology employed to develop the \textit{Psycho Gundam} system, an Electroencephalography (EEG)-based real-time robotic control system using deep learning. The system leverages the EMOTIV+ EEG cap to capture brainwave signals and translate them into robotic control actions through a series of preprocessing, feature extraction, and deep learning-based classification steps. The goal is to enable real-time, accurate control of a robotic system by leveraging cognitive signals from human subjects.

\subsection{EEG Data Acquisition}

\subsubsection{Equipment}
EEG data was collected using the EMOTIV+ EEG cap, a non-invasive, wireless EEG headset. The cap is equipped with multiple electrodes placed according to the International 10-20 system to record electrical activity in the brain. The EEG signals were sampled at a frequency of 128 Hz for optimal resolution, capturing various frequency bands of interest such as delta (1-4 Hz), theta (4-8 Hz), alpha (8-12 Hz), beta (12-30 Hz), and gamma (>30 Hz).

\subsubsection{Subjects}
One participant, aged 27 years, left-handed, was selected for the study. The participants were trained on the intended mental tasks, such as imagining left or right hand movements, or focusing on specific visual targets to invoke corresponding EEG patterns.

\subsubsection{Experimental Setup}
Each subject wore the EMOTIV+ EEG cap while seated in front of the robotic system. During the experiment, subjects were instructed to perform mental tasks corresponding to specific robotic actions, such as moving the robot arm forward, backward, left, or right. Data was recorded in sessions lasting 10 minutes, with rest intervals to prevent mental fatigue.

\subsection{Signal Processing}

\subsubsection{Standard Procedure}

Before analysis, raw EEG data is typically subjected to the following preprocessing steps to remove noise and artifacts:

\begin{itemize} \item \textbf{Band-pass filtering}: A band-pass filter between 1 Hz and 50 Hz is usually applied to eliminate powerline noise and non-relevant low/high-frequency components. \item \textbf{Artifact removal}: Independent Component Analysis (ICA) is used to remove artifacts such as eye blinks, muscle movements, and other physiological noise. \item \textbf{Normalization}: The signals are normalized to ensure consistency across different sessions and participants. \end{itemize}

\subsubsection{Current Practical Approach}

Since the latest free version of the EMOTIV software no longer supports exporting raw EEG data, we have switched to recording the screen to capture raw EEG data as video files, which now serve as the primary data source for analysis.
\begin{itemize}
    \item \textbf{Denoising} After converting the screen-recorded EEG signals into images, we first manually remove frames with noise, such as those containing unintended windows from accidental mouse clicks or frames where unrelated visuals obstruct the EEG recording screen.
\end{itemize}

\subsection{Feature Extraction}

\subsubsection{If Viewing EEG as Time-Series Data}
After preprocessing, key features were extracted from the EEG signals. Time-domain, frequency-domain, and time-frequency domain features were considered:
\begin{itemize}
    \item \textbf{Power Spectral Density (PSD)}: Extracted using Welch's method to capture frequency band power.
    \item \textbf{Time-domain features}: Statistical measures such as mean, variance, and skewness were calculated over sliding windows.
    \item \textbf{Wavelet Transform}: Applied to capture non-stationary EEG signal properties, particularly in the alpha and beta bands.
\end{itemize}

\subsubsection{If Viewing EEG as Images from Video}

In this approach, EEG data is treated as sequences of images extracted from recorded video. A neural network is used for feature extraction, allowing for spatial-temporal analysis and leveraging deep learning methods typically applied in computer vision to capture complex patterns and dynamics in the data.

\subsection{Deep Learning Model for EEG-to-Robot Mapping}

\subsubsection{Model Architecture}
A deep learning model was developed to map EEG signals to corresponding robot control commands. The architecture consists of:
\begin{itemize}
    \item \textbf{Input Layer}: The processed EEG feature vectors serve as input to the model.
    \item \textbf{Pre-trained Vision Transformer (ViT) \cite{alexey2020image}}: The EEG feature vectors are fed into a pre-trained Vision Transformer model, which leverages self-attention mechanisms to capture both spatial and temporal dependencies within the EEG data, providing a comprehensive understanding of the signal patterns.
    \item \textbf{Fully Connected Layer}: The output embeddings from the ViT model are passed through fully connected layers to map the transformed features to robot control actions (e.g., move left, right, forward, backward).
    \item \textbf{Output Layer}: The final layer provides categorical outputs corresponding to discrete robotic control commands.
\end{itemize}

\subsubsection{Training Procedure}
The model was trained on a dataset comprising EEG data paired with labeled control actions:
\begin{itemize}
    \item \textbf{Loss Function}: Cross-entropy loss was utilized to optimize classification accuracy.
    \item \textbf{Optimizer}: The Adam optimizer with a learning rate of 0.001 was used to train the model.
    \item \textbf{Training/Validation Split}: To prevent overfitting, the dataset was divided into 80\% training and 20\% validation sets.
\end{itemize}

\subsection{Robotic System Integration}

\subsubsection{Robotic Platform}
The EEG-based control system is integrated with a robotic platform designed for multi-axis movement (Figure 2), the more details are introduced in \cite{gundam}. The control interface receives EEG-based commands in real-time via a USB connection between the robot controller and the computing system. To minimize excessive mechanical jitter, EEG-based commands are processed through a specially designed low-pass filter. The overall motion design of the robot is inspired by classic fighting games. To enhance the real-time control experience like playing the fighting games, the low-pass filter includes an integrator, which accumulates non-responsive signals and triggers specific actions once a threshold is reached.

\subsubsection{Robot Motion Control}
The robot control system simplifies EEG signals into five types: forward, defense, punch, heavy punch and kick. It also supports some classic fighting game combos, though the operational complexity has been simplified due to signal limitations(Table 1). Each of these actions activates a pre-defined motion control trajectory. The motors themselves provide angle feedback control, so the controller is only responsible for transmitting the angle signals.

\begin{figure}
  \begin{minipage}[b]{.45\linewidth}
    \centering
    \includegraphics[width=0.7\linewidth]{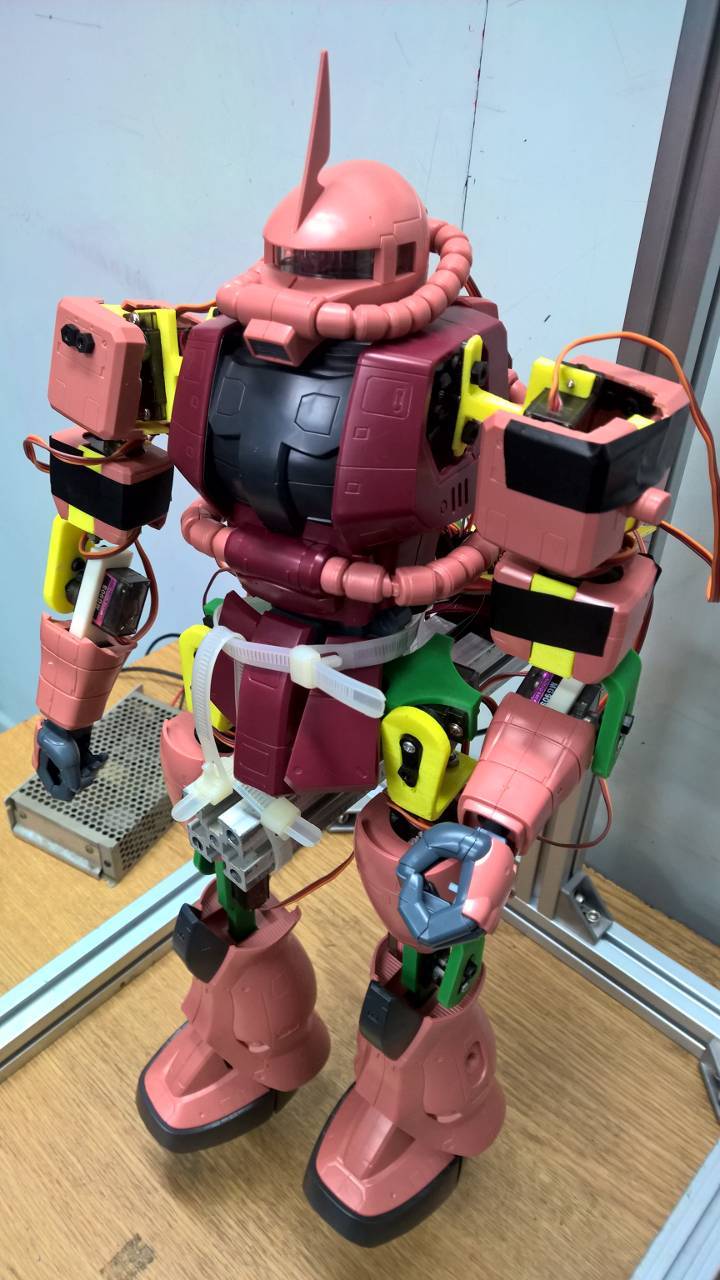}
    \captionof{figure}{A robot made by hollowing out a MEGA size gundam and putting a motor in it.}
  \end{minipage}\hfill
  \begin{minipage}[b]{.45\linewidth}
    \centering
    \begin{tabular}{ *{3}{c} }
        \hline
        number & move & operate \\
        \hline
        1 & defense & left pedal \\
        \hline
        2 & forward & right pedal \\
        \hline
        3 & punch & left Handle \\
        \hline
        4 & heavy punch & both handle \\
        \hline
        5 & kick & right Handle \\
        \hline
        6 & punch combo & forward + punch\\
        \hline
        7 & Uppercut & forward + heavy punch\\
        \hline
        8 & kick combo & forward + kick\\
        \hline
        9 & Hadoken & punch + heavy punch\\
        \hline
    \end{tabular}
    \captionof{table}{A mapping table of robot moves.}
  \end{minipage}
\end{figure}

\section{Result and Discussion}
By using a pre-trained Vision Transformer (ViT), we achieved an overall classification accuracy of 57.41\% on our custom dataset designed to simulate the Psycho Gundam cockpit environment. The lower accuracy may stem from the fact that our raw EEG data did not undergo traditional signal filtering for time-series processing, as typically applied in EEG analysis. Nevertheless, as a pioneering Motor Imagery (MI) dataset, this represents a novel approach and serves as an important first step in exploring this unique application. Due to the high variability of brainwave patterns across different individuals and sessions, achieving smooth and reliable mapping to control outputs remains a significant challenge. EEG signals are inherently noisy and vary with factors like mental state and external distractions, which complicates the model’s ability to generalize. This variability underscores the need for more advanced methods in feature extraction and model training to ensure stable, responsive performance in real-time applications.

\section{Conclusion}
In this work, we introduced the first Motor Imagery (MI) dataset based on a Gundam cockpit simulation and developed a real-time EEG-controlled robotic system. This pioneering approach demonstrates the potential for integrating EEG-based control into complex robotic applications, paving the way for future advancements in brain-machine interfaces. Our results highlight both the challenges and opportunities presented by raw EEG data in real-time control scenarios, especially within unique, non-traditional contexts. Additionally, recent advancements in quantum computing for EEG processing and AI \cite{chen2024quantum, chen2024qeegnet} provide promising avenues for further exploration. The integration of quantum machine learning could bring us closer to transforming the science fiction of anime into reality. Future work will focus on refining the signal processing pipeline and enhancing model accuracy, aiming to improve the responsiveness and reliability of EEG-driven control systems. We hope this research inspires further exploration into novel applications of EEG for immersive, interactive robotic experiences.


\bibliographystyle{unsrtnat}
\bibliography{references}  

\begin{thebibliography}{10}
\providecommand{\natexlab}[1]{#1}
\providecommand{\url}[1]{\texttt{#1}}
\expandafter\ifx\csname urlstyle\endcsname\relax
  \providecommand{\doi}[1]{doi: #1}\else
  \providecommand{\doi}{doi: \begingroup \urlstyle{rm}\Url}\fi

\bibitem[sai Wu(2023)]{gundam}
Wei-Sheng~Wang sai Wu.
\newblock Realization of gundam ’s key technologies and feasibility assessment.
\newblock \emph{U-ACG}, 2023.

\bibitem[An et~al.(2023)An, Wong, and Ling]{an2023eeg}
Yang An, Johnny~KW Wong, and Sai~Ho Ling.
\newblock An eeg-based brain-computer interface for real-time multi-task robotic control.
\newblock In \emph{2023 45th Annual International Conference of the IEEE Engineering in Medicine \& Biology Society (EMBC)}, pages 1--4. IEEE, 2023.

\bibitem[Xu et~al.(2022)Xu, Li, Liu, Zhang, Miao, Xu, and Song]{xu2022continuous}
Baoguo Xu, Wenlong Li, Deping Liu, Kun Zhang, Minmin Miao, Guozheng Xu, and Aiguo Song.
\newblock Continuous hybrid bci control for robotic arm using noninvasive electroencephalogram, computer vision, and eye tracking.
\newblock \emph{Mathematics}, 10\penalty0 (4):\penalty0 618, 2022.

\bibitem[Wolpaw et~al.(2002)Wolpaw, Birbaumer, McFarland, Pfurtscheller, and Vaughan]{wolpaw2002brain}
Jonathan~R Wolpaw, Niels Birbaumer, Dennis~J McFarland, Gert Pfurtscheller, and Theresa~M Vaughan.
\newblock Brain--computer interfaces for communication and control.
\newblock \emph{Clinical neurophysiology}, 113\penalty0 (6):\penalty0 767--791, 2002.

\bibitem[Chen and Wei(2024)]{chen2024mind}
Chi-Sheng Chen and Chun-Shu Wei.
\newblock Mind's eye: Image recognition by eeg via multimodal similarity-keeping contrastive learning.
\newblock \emph{arXiv preprint arXiv:2406.16910}, 2024.

\bibitem[Chen(2024)]{chen2024necomimi}
Chi-Sheng Chen.
\newblock Necomimi: Neural-cognitive multimodal eeg-informed image generation with diffusion models.
\newblock \emph{arXiv preprint arXiv:2410.00712}, 2024.

\bibitem[Lawhern et~al.(2018)Lawhern, Solon, Waytowich, Gordon, Hung, and Lance]{lawhern2018eegnet}
Vernon~J Lawhern, Amelia~J Solon, Nicholas~R Waytowich, Stephen~M Gordon, Chou~P Hung, and Brent~J Lance.
\newblock Eegnet: a compact convolutional neural network for eeg-based brain--computer interfaces.
\newblock \emph{Journal of neural engineering}, 15\penalty0 (5):\penalty0 056013, 2018.

\bibitem[Alexey(2020)]{alexey2020image}
Dosovitskiy Alexey.
\newblock An image is worth 16x16 words: Transformers for image recognition at scale.
\newblock \emph{arXiv preprint arXiv: 2010.11929}, 2020.

\bibitem[Chen et~al.(2024{\natexlab{a}})Chen, Tsai, and Huang]{chen2024quantum}
Chi-Sheng Chen, Aidan Hung-Wen Tsai, and Sheng-Chieh Huang.
\newblock Quantum multimodal contrastive learning framework.
\newblock \emph{arXiv preprint arXiv:2408.13919}, 2024{\natexlab{a}}.

\bibitem[Chen et~al.(2024{\natexlab{b}})Chen, Chen, Tsai, and Wei]{chen2024qeegnet}
Chi-Sheng Chen, Samuel Yen-Chi Chen, Aidan Hung-Wen Tsai, and Chun-Shu Wei.
\newblock Qeegnet: Quantum machine learning for enhanced electroencephalography encoding.
\newblock \emph{arXiv preprint arXiv:2407.19214}, 2024{\natexlab{b}}.

\end{thebibliography}





\newpage
\appendix
\section{Appendix}
\subsection{More experiment details}
The project code is available at \url{https://github.com/ChiShengChen/PSYCHO_GUNDAM}.

\begin{table}[h]
    \centering
    \begin{tabular}{ccc}
      Category & Label &  Number of Images  \\
      \hline
      Pull Forward with Both Hands   & 0  & 18052\\
      Left Leg on the Pedal & 1  & 18020\\
      Push with 2 Hands  & 2 & 18024 \\
      Right Leg on the Pedal  & 3 & 18022 \\
      Push Upward with Both Hands& 4 & 18050\\
      \hline
    \end{tabular}
    \caption{The details of the whole dataset.}
    \label{tab:my_label}
\end{table}

\subsection{Update}
This paper has been accepted by 2024 13th International Symposium on ACG Culture and Technology Studies (\url{https://u-acg.com/archives/29015}).

\end{document}